 \pdfoutput=1

 \documentclass[a4paper, 10 pt]{article}  % Comment this line out if you need a4paper

\usepackage{graphicx} %use graph format
\usepackage{epstopdf}
\usepackage{mathptmx} % assumes new font selection scheme installed
\usepackage{times} % assumes new font selection scheme installed
\usepackage{amsmath} % assumes amsmath package installed
\usepackage{amssymb}  % assumes amsmath package installed
\usepackage{cite}
\usepackage{algorithm}  
\usepackage{algorithmicx}  
\usepackage{algpseudocode}

\title{\LARGE \bf
Autonomous Underwater Vehicle-Manipulator Systems Path Planning with RRTAUVMS Algorithm 
}

\author{Xiaoxu Cao$^{1,2,3}$, Linyi Gu$^{1}$, JunChen Mu$^1$, Qian Zhang$^{1}$, Qi Song$^{1}$ \\ Chunxiao Liu$^{2}$, Cong Qiu$^{2}$
\thanks{$^{1}$  Zhejiang University, State Key Lab of Fluid Power \&  Mechatronic Systems, Hangzhou, CN
{\tt\small caoxiaoxu@zju.edu.cn}
        } 
\thanks{$^{2}$  Shenzhen Sensetime Technology CO.,LTD 
        }  
\thanks{$^{3}$	Shenzhen Institutes of Advanced Technology, Chinese Academy of Sciences
        }
}

\begin{document}

\maketitle
\thispagestyle{empty}
\pagestyle{empty}

%%%%%%%%%%%%%%%%%%%%%%%%%%%%%%%%%%%%%%%%%%%%%%%%%%%%%%%%%%%%%%%%%%%%%%%%%%%%%%%%
\begin{abstract}

Autonomous Underwater Vehicle-Manipulator systems (AUVMS) is a new tool for ocean exploration, the AUVMS path planning problem is addressed in this paper. AUVMS is a high dimension system with a large difference in inertia distribution, also it works in a complex environment with obstacles.  By integrating the rapidly-exploring random tree(RRT) algorithm with the AUVMS kinematics model, the proposed RRTAUVMS algorithm could randomly sample in the configuration space(C-Space), and also grow the tree directly towards the workspace goal in the task space. The RRTAUVMS can also deal with the redundant mapping of workspace planning goal and configuration space goal. Compared with the traditional RRT algorithm, the efficiency of the AUVMS path planning can be significantly improved.

\end{abstract}

%%%%%%%%%%%%%%%%%%%%%%%%%%%%%%%%%%%%%%%%%%%%%%%%%%%%%%%%%%%%%%%%%%%%%%%%%%%%%%%%
\section{INTRODUCTION}
AUVMS is the new tool for ocean exploration, some prototypes have been proposed to 
evaluate the work performance of underwater intervention tasks\cite{yuh_design_1998,sanz_trident:_2010,simetti_floating_2014,Maurelli2016}.  Although the traditional working class remotely operated vehicles(ROV)-Manipulator Systems have already been deployed in many underwater tasks with proven techniques, the ROVMS(ROV-Manipulator Systems) requires many expensive auxiliary systems, such as the Tether Management System and Dynamic Position system to reduce the risk of umbilical cable broken\cite{ridao_intervention_2014}. Experienced onboard operators are required when conducting the underwater tasks, however, the fatal accident like ROV intertwining with underwater facilities still can't be totally avoided. Without umbilical cable, the AUVuMS  pose obvious advantages in long distance inspection task, and manipulation work in the multi-obstacles environment, however, the complex working environment also makes the coordinate motion planning of AUVMS a core research topic. 

Motion planning technology is one of the key technologies to realize the autonomous operation. For autonomous robots, the motion planning is divided into two parts, path planning, mainly to search the collision-free paths in the global scope; trajectory planning, solving the problems of motion allocation and freedom redundancy processing of highly redundant systems. 

The pseudo-inverse method is one of the widely used solutions for redundant systems, the optimal vehicle/joint velocities in configuration space can be obtained, this also applies to the weighted pseudo-inverse method, which will lead to the minimization of weighted vehicle/joint velocities norm\cite{DEWhitney1969,klein_review_1983}.

The task priority redundancy resolution techniques are proposed by Gianluca Antonelli\cite{GAntonelli1998Task}, this method can guarantee the end effector(eef) trajectory tracking while at the same time, achieves a second task such as the optimal energy consumption or robot manipulability. In paper\cite{GAntonelli2009Prioritized} the task priority method is used to solve the output saturation problem. After that, further researches are proposed in \cite{antonelli_fuzzy_redundancy_2003,antonelli_fuzzy_approch_2003} to merge  the fuzzy algorithm with task priority approach. The fuzzy technique is used to coordinate the  multiple secondary tasks, the multi-task motion planning  can be achieved with this method.

Tremendous research effort has been devoted to the robot motion planning. The randomized sampling based algorithms have been widely used for robots high-dimensional motion planning. In 1998, the Rapidly-exploring Random Tree (RRT) algorithm was proposed\cite{kavraki_probabilistic_1996,Lavalle1998Rapidly}. The RRT algorithm can efficiently search the path in the high-dimensional C-space, also the resulting path can guarantee no violation with obstacles. The RRT algorithm and its variations have been used in the underwater robots planning scenarios. The motion planning of AUV in underwater environment has been studied based on RRT algorithm\cite{hernandez_online_2015,carreras_online_2016}, the optimal rapidly-exploring random tree (RRT*) is extended  using concepts of anytime algorithms and lazy collision evaluation. The motion planning of underwater vehicle manipulator systems for autonomous underwater inspection operations is investigated\cite{Sotiropoulos2015Rapid}. The collision avoidance, approximation of the given task curve and critical optimization criteria are considered in the planning algorithm. The planner can deal with the unknown obstacles inside the workspace while executing the task. The RRT-connect algorithm is implemented in GIRONA500 UVMS motion planning\cite{youakim_moveit_2017}, RRT-connect works by growing two separated trees of paths rooted at the start and goal poses. The trees explore the space around them by progressing toward one another using a simple greedy heuristic until they join together, it's an efficient path searching method, and it is suitable for high-dimensional C-space.

The RRT algorithm is also used in redundant robot system planning. For redundant system, solving the inverse kinematics(IK) is challenging, the numerical approximation method provides a way for IK solution, however, this method has high computational complexity, and sometimes can't converge to a valid solution. So in\cite{bertram_integrated_2006}, the heuristic workspace metric which measures the manipulator's distance to goal is used to guide the search in C-Space, the IK problem is avoided here. In \cite{Weghe2007Randomized}, the Jacobian transpose is used to guide the RRT tree growing direction, the modified RRT algorithm is implemented in the 7 degree-of-freedom robotic arm. The IK solution is not required using the proposed algorithm, the Jacobian inverse solution is also avoided here. However, the Jacobian transpose based solution can not take the joint inertia difference into account, also the joint limits constrained is not considered. So the algorithm in paper\cite{Weghe2007Randomized} doesn't apply to the AUVMS system, which is composed of two subsystems with a large inertia difference, and the manipulator joint limits constrain is considered.

In this paper, the AUVMS kinematics constrains are integrated with the RRT algorithm to solve the AUVMS path planning problem. The core idea can be summarized as: Using the pseudo-inverse Jacobian matrix to guide the RRT tree exploration in the workspace, while randomly sampling in C-space. So in this paper, we will investigate how to achieve the whole body motion planning for a high dexterous system with large inertia difference. This paper is organized as follows. In the next section, a brief description of AUVMS kinematics model are given.

\section{PROBLEM FORMULATION}

\subsection{AUVMS Kinematics Model}
The system configuration of AUVMS is shown in Fig\ref{fig:AUVMSsystem}, the manipulator with 4 revolute joints is mounted on the left side of AUV. The body-fixed frame is built on the AUV center of gravity, the eef frame is built on the end tip of AUV, the earth-fixed reference frame aligns with the body-fixed frame. The eef position \& pose can be expressed as:

\begin{figure}[H]
      \centering
      \includegraphics[width=0.3\textwidth]{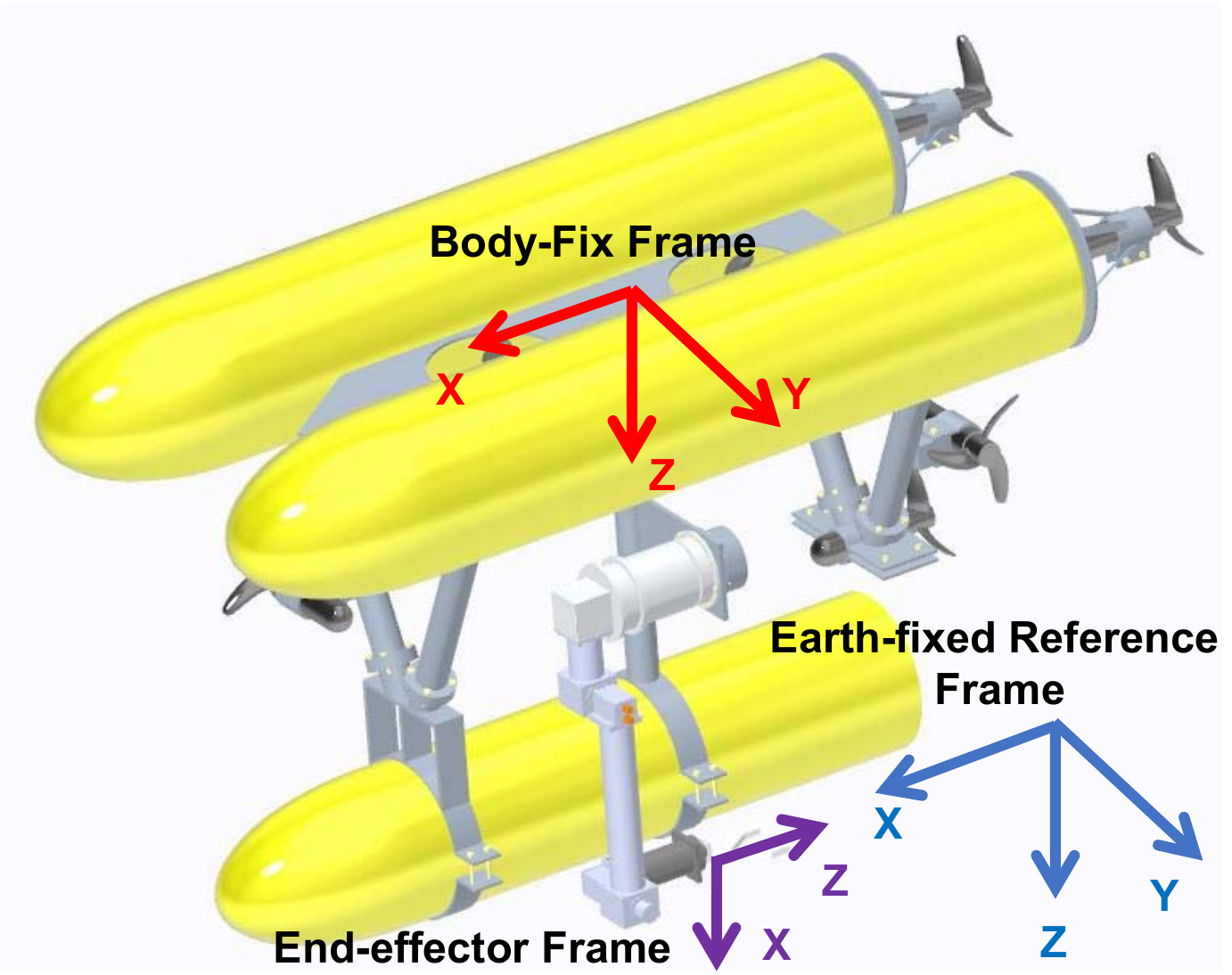}
      \caption{AUVMS system configuration and frame definition}
      \label{fig:AUVMSsystem}
   \end{figure}

First, the relationship of body-fixed velocity of AUV and the velocity in earth-fixed frame of AUV can be expressed as:

\begin{equation}
\dot{\eta}=
{\left[ \begin{array}{cc}
  J_{v1}(\eta_2)& 0                \\
  0            & J_{v2}(\eta_2) 
  \end{array} 
  \right ]}\nu
\label{eq:AUV_kinematics}
\end{equation}

Where $J_{v1}, J_{v2}$ are the Jacobian matrixs of AUV,

\begin{equation}
\label{eq:jv1}
J_{v1}={
  \left[ \begin{array}{ccc}
  c{\psi}c{\theta}  &  -s{\psi}c{\phi}+c{\psi}s{\theta}s{\phi} &  s{\psi}s{\phi}+c{\psi}s{\theta}c{\phi}\\
  s{\psi}c{\theta} & c{\psi}c{\phi}+s{\psi}s{\theta}s{\phi}  &-c{\psi}s{\phi}+s{\psi}s{\theta}c{\phi}  \\
  -s{\theta} & s{\phi}c{\theta}  & c{\phi}c{\theta}
  \end{array} 
  \right ]}
\end{equation}

\begin{equation}
\label{eq:jv2}
J_{v2}=\frac{1}{c{\theta}}{
  \left[ \begin{array}{ccc}
  1 & s{\phi}s{\theta}   & c{\phi}s{\theta}    \\
  0 & c{\phi}c{\theta} &-c{\theta}s{\phi} \\
  0 & s{\phi}         &c{\phi} 
  \end{array} 
  \right ]}
\end{equation}

Where $s\bullet  =\sin{\bullet}, c\bullet = \cos{\bullet}$.

For manipulator system, the kinematics can be derived using the D-H method.

The position and pose of eef in the body-fixed frame can be defined as:

\begin{equation}
\eta_{ve}=\begin{bmatrix}
\eta^T_{ve1} & \eta^T_{ve2}
\end{bmatrix}^T
\end{equation}

Where $\eta_{ve1}=[x_{ve} , y_{ve} ,z_{ve} ]^T$ is the position of eef in body-fixed frame, $\eta_{ve2}= [\phi_{ve} , \theta_{ve} , \psi_{ve}]^T $ is the pose vector of eef in body-fixed frame.

Manipulator Jacobian matrix describes the relationship of joint space velocity and eef velocity in body-fixed frame.

\begin{equation}
\dot{\eta}_{ve}=J_m\dot{q}=\begin{bmatrix}
J_{mp}(Q) \\
J_{mo}(Q)
\end{bmatrix} \dot{Q}
\label{eq:arm_kinematics}
\end{equation}

Where, $Q=[q_1,q_2,q_3,q_4]^T$ is the joint position vector, $\dot{Q}$ is the joint velocity; $\dot{\eta}_{ve} \in  \mathbb{R}^6$is the manipulator eef velocity in body-fixed frame, $J_m\in \mathbb{R}^{6\times 4}$ is the manipulator Jacobian matrix.

$S(\omega)$ donates the skew-symmetric matrix, $\omega=[\omega_x, \omega_y, \omega_z]^T$, then $S(\omega)$ is
\begin{equation}
\label{skew_symmetric_operator}
S(\omega)=\begin{bmatrix}
0 & -\omega_z & \omega_y \\
\omega_z & 0 & -\omega_x \\
-\omega_y & \omega_x & 0 
\end{bmatrix}
\end{equation}

Jacobian matrix describe the relationship of configuration space velocity ($\zeta=[\nu_1, \nu_2, \dot{Q} ]^T\in \mathbb{R}^8 $)and the eef speed in working space($\dot{\eta}_{ie} =[\eta^T_{ie1} \eta^T_{ie2} ]^T \in  \mathbb{R}^6 $), $\eta_{ie1}=[x_{ie} , y_{ie} , z_{ie} ]^T$ is the position vector in reference frame, $\eta_{ie2}=[\phi_{ie} , \theta_{ie} , \psi_{ie}]^T$ is the pose vector.

  \begin{equation}
  \dot{\eta}_{ie}=J_{auvms}\zeta
  \label{jauvms}
  \end{equation}

  The Jacobian matrix of the whole system can be expressed as:

  \begin{equation}
  J_{auvms}=\begin{bmatrix}
J_{v1}(\eta_2) & -S(J_{v1}(\eta_2)\eta_{ve1})J_{v2}(\eta_2)  & J_{v1}(\eta_2)J_{mp} \\
0              &  J_{v2}(\eta_2)                            &   J_{v1}(\eta_2)J_{mo}
\end{bmatrix}
  \end{equation}

\subsection{Weighted Pseudo-Inverse Solution}

As the system is redundant, solution of Eq.\ref{jauvms} can not be directly obtained by matrix inverse. So the pseudo-inverse is derived by the following steps.

Define the quadratic cost function:

    \begin{equation}
  E=\frac{1}{2}\zeta^T W \zeta
  \label{cost_weighted_pesu}
  \end{equation}
  
  Where $W$ is the weight for each joint in C-space. Then optimal solution which can minimize the cost function Eq.\ref{cost_weighted_pesu} can be derived as:
  
  \begin{equation}
  \label{pesudoweighted_jacob}
  \zeta =W^{-1}J^{T}_{auvms}(J_{auvms} W^{-1} J^T_{auvms})^{-1} \dot{\eta}_{ie}
  \end{equation}

The weighted pseudo-inverse can be expressed as:
  \begin{equation}
  \label{pesudoweighted_jacob_inverse}
  J^{\dagger}_{auvms}=W^{-1}J^{T}_{auvms}(J_{auvms} W^{-1} J^T_{auvms})^{-1}
  \end{equation}

 Consider the joint position constrain problem, in the motion planning, to avoid the invalid planning results, for example the joint position out of limit, the motion planning considering the joint constrain are proposed here.

 Design the cost function as:
    
    \begin{equation}
    H(q)=\sum_{i=1}^{4} \frac{(q_{i,max}-q_{i,min})}{C_i (q_{i.max}-q_i)(q_i-q_{i,min})} 
    \label{chapter3:cost_function}
    \end{equation}
    
    For joint $i$, $C_i>0$, $q_i$ is the joint position, and $q_i \in (q_{i,max},q_{i,min})$, the partial derivative of $H(q)$ is:
    
    \begin{equation}
    \frac{\partial  H(q) }{\partial q_i}=\frac{1}{C_i} \frac{(q_{i,max}-q_{i,min})(2q_i-q_{i,max}-q_{i,min})}{(q_{i.max}-q_i)^2(q_i-q_{i,min})^2} 
    \label{chapter3:partial_cost}
    \end{equation}
    
    so the weight for joint $i$ is
    
    \begin{equation}
    w_i=1+\left \|  \frac{\partial  H(q) }{\partial q_i} \right \|
    \label{chapter3:weight_matrix}
    \end{equation}
    
   so the weight matrix for manipulator is:
    
    \begin{equation}
    W_m=\mathrm{diag}[w_1,w_2,w_3,w_4]
    \end{equation}

Then the whole system weight matrix can be defined as:

  \begin{equation}
    W=\mathrm{diag}[w_x,w_y,w_z,w_r,w_1,w_2,w_3,w_4]
    \end{equation}

So far, the AUVMS kinematic model and the weight matrix has been obtained, they will play a important role in the new path planning algorithm.

\section{RRTAUVMS ALGORITHM}

The traditional RRT algorithm will sample in the C-space uniformly, also the start state and goal state are in C-space are needed. As we explained before, the system is constrained by the kinematic model, so sampling in the workspace and then map the extension into C-space, that will be more efficient. Also, the goal state are normally given in workspace, such as a position in the workspace, the goal state should be mapped into C-space at first for the traditional RRT, however for redundant system, the mapping will lead to a variety of optimization problems. Here the RRTAUVMS algorithm is proposed the for the motion planning of AUVMS.

\begin{algorithm}
  \caption{RRTAUVMS Algorithm}  
  \label{RRTAUVMS_Algorithm}
  \begin{algorithmic}[1]   
    \Require Map, Start, goal, Step length in CSpace,Step length in WSpace
    \Ensure  

    \Function {RRTAUVMS}{start, goal, K, CStep, WStep}  
    \State Tree.init($x_{init}$)    
    \While{ DistancetoGoal$()>$Threshold}
    \For{ i=1 to K}
    \State p=rand
    \If{$p<p_g$}    
    \State Tree=ExtendRandomly(start, goal, CStep,Tree)
    \Else 
    \State Tree =ExtendToGoal(start, goal, WStep,Tree)
    \EndIf
    \EndFor
    \EndWhile   
    \EndFunction  
    
      \Function {ExtendToGoal}{start, goal, WStep,Tree}  
      \State $x_{nearest}$=ClosestNodetoGoal(Tree)
      
      \State Jauvms\_pesudo=JacobianCal($x_{nearest}$)
      
      \State  $\Delta_x$=WorkspaceStep($x_{nearest}$, workspacegoal, WStep)
      \State  $\Delta_q$= Jauvms\_pesudo $\Delta_x$
      \State $x_{new}$=$x_{nearest}$+$\Delta_q$
        \If{ObstacleFree($x_{new}$)}
        \State Tree.add\_vertex($x_{new}$)
        \EndIf
        
        \State Return Tree
            
      \EndFunction  
  \end{algorithmic}  
\end{algorithm}

\begin{algorithm}
  \caption{Path Smooth Algorithm}  
  \label{SmoothPath}
  \begin{algorithmic}[1] 
    \Require Map, Raw path
    \Ensure  
    \Function {SmoothPath}(PathList,map)  
    \State n=length(PathList)
    \State StartIndex=1, EndIndex=n, SmoothList=PathList(1);
    
    \While{StartIndex $<$ n}
    \While {StartIndex $<$ EndIndex}
    \If {isEdgeBelongsFreeSpace(PathList(StartIndex),PathList(EndIndex), map)}
    \State SmoothPaht.add(PathList(EndList))
    \State StartIndex=EndIndex
    \Else
    \State EndIndex=EndIndex-1
    \EndIf
    \EndWhile
    
    \EndWhile

    \EndFunction  
  
  \end{algorithmic}  
\end{algorithm}  

\begin{figure}[H]
      \centering
      \includegraphics[width=0.4\textwidth]{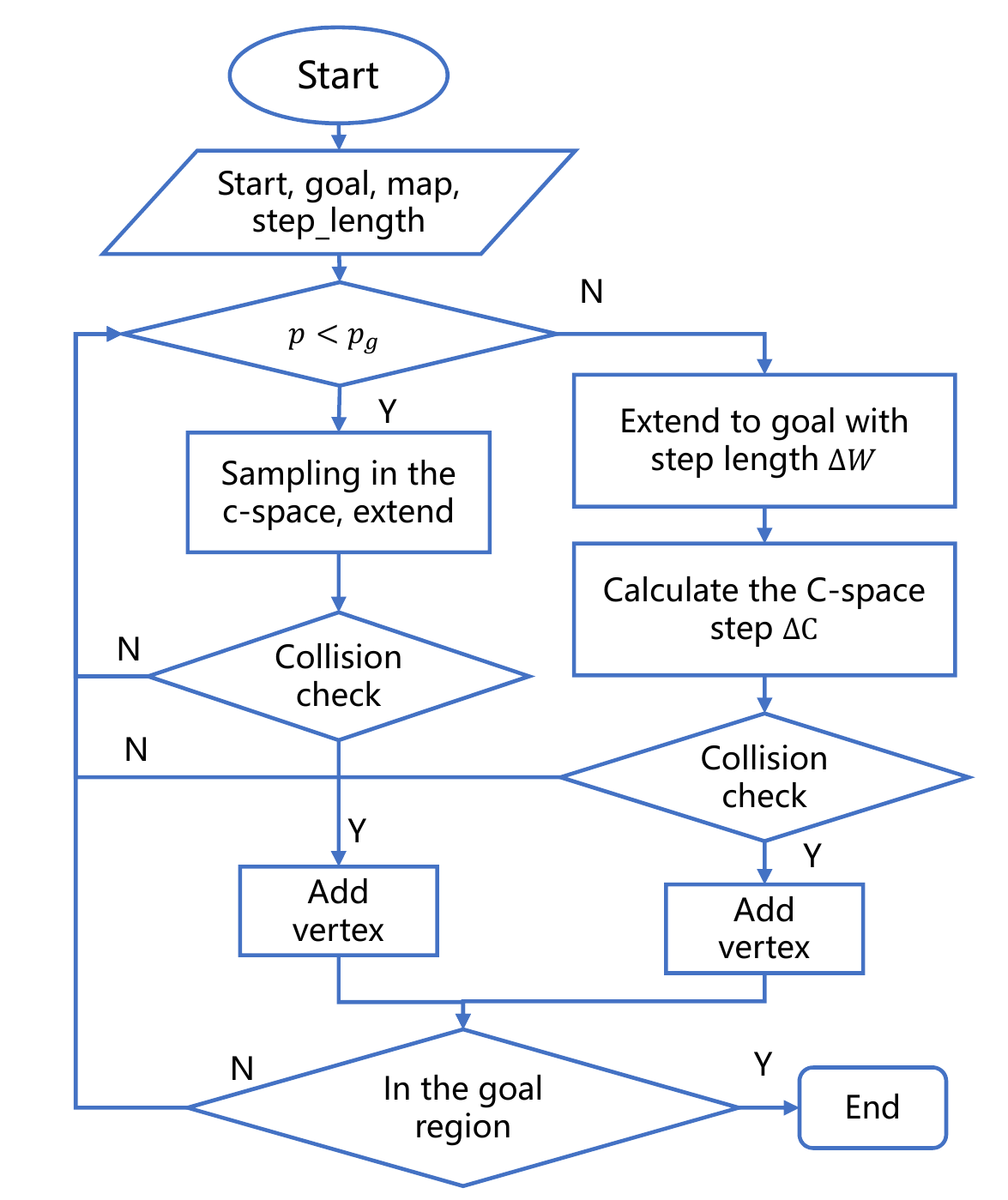}
      \caption{RRTAUVMS algorithm flow chart}
      \label{fig:RRTAUVMS_flowchart}
\end{figure}

The main algorithm is shown in Algorithm.\ref{RRTAUVMS_Algorithm}. In the pseudo code, the \textbf{RRTAUVMS} is the main function, the start state, goal state, maximum iteration steps K, C-space step: CStep and work space step WStep are the variables  of the function. At the beginning of the algorithm, a Tree is created with only one node, the node corresponding to the start state. Then the algorithm enter the \textbf{while} loop, for every iteration step, the algorithm will generate a random number $p \in [0,1]$, if $p$ is larger than the possibility threshold $p_g$, then the tree will extend to the workspace goal with function \textbf{ExtendToGoal}, or the tree will extend randomly with function \textbf{ExtendRandomly}. The function \textbf{ExtendRandomly} is just the same as the traditional RRT, it will sample in the feasible C-space, while the \textbf{ExtendToGoal}, will sample in the work space. As shown in the pseudo code, the first step  of function \textbf{ExtendToGoal} is find the nearest node to the goal, which is expressed as $x_{nearest}$. Then the Jacobian matrix of the system state $x_{nearest}$ and pseudo-inverse Jacobian matrix are obtained. The work space extension vector $\Delta x$ can be calculate from \textbf{WorkspaceStep} function,  the vector $\Delta x$ starts from $x_{nearest}$  and ends in workspace goal. Because the step extension can be treat as an instantaneous  movement direction, the Eq.\ref{pesudoweighted_jacob} can be applied to calculate the step extension $\Delta q$ in the configuration space, then the new state $x_{new}$ is obtained. If the new state is collision-free, then it's added to the tree. Here the selection of Jacobian inverse is important, when compute the new state  $x_{new}$, the joints angle may be out of the limits, however, if the weighted Jacobian pseudo-inverse Eq.\ref{pesudoweighted_jacob_inverse} is used, the new state will not violate the joint  limits constrain. The \textbf{RRTAUVMS} function working mechanism is shown in Fig.\ref{fig:RRTAUVMS_flowchart}.

After the path searching procedure, a raw path with a lot of nodes can be obtained. The raw path is not suitable for robot path following as a lot of sharp peak are observed in the path. The Algorithm.\ref{SmoothPath} is used to get a smooth path with little nodes. It works as following, in the raw path list, connect the start node(StartIndex) and then end node (EndIndex), then check collision. If it's not collision free, check the edge between (EndIndex-1) node and start node(StartIndex), if it's collision free, the edge is kept in the smooth list, and make the StartIndex equals the EndIndex, the algorithm will iterate until the StartIndex equals to the length of the raw path. 

\section{SIMULATION RESULTS}

Considering the real working conditions, the simulation is designed as following. We will command the end-effector of manipulator to move toward the goal position, get around the obstacles on the way. 

The goal position is set to be $[x_g,y_g,z_g]$=[4, 4, 4]m; and the goal ball region is 0.3m; the CSpace step CStep is [0.1, 0.1 0.1 0.08, 0.05, 0.05, 0.05, 0.05]; the working space step WStep is: [0.2 0.2 0.2 0.05 0.05 0.05]; The start state in CSpace is $[x_v,y_v,z_v,\psi ,q_1,q_2,q_3,q_4]=[0,0,0,0,0,0,0,\pi ]$, the corresponding position and pose in working space is $[x,y,z,\varphi ,\theta ,\psi ]=[-0.1176,0.1898,-0.4120,0,-1.5708,0]$; the obstacles position and radius is:$P_{obs1}=[2, 2, 3]^T, r_{obs1}=0.3$; $P_{obs2}=[1, 1, 1]^T, r_{obs2}=0.2$;  $P_{obs3}=[1, 3, 2]^T, r_{obs3}=0.3$;

To evaluate the performance of of RRTAUVMS algorithm, contrast experiments are conducted to compare the efficiency of traditional RRT algorithm and RRTAUVMS algorithm. The single obstacle and multi obstacles environments are evaluated. In the simulation experiments, the program will capture the time when the searching is completed.  The simulation platform is MATLAB/Simulink 2015a running on PC with i5-3320M 2.6GHz CPU.

\begin{figure}[htbp]
	\centering
	\includegraphics[scale=0.4]{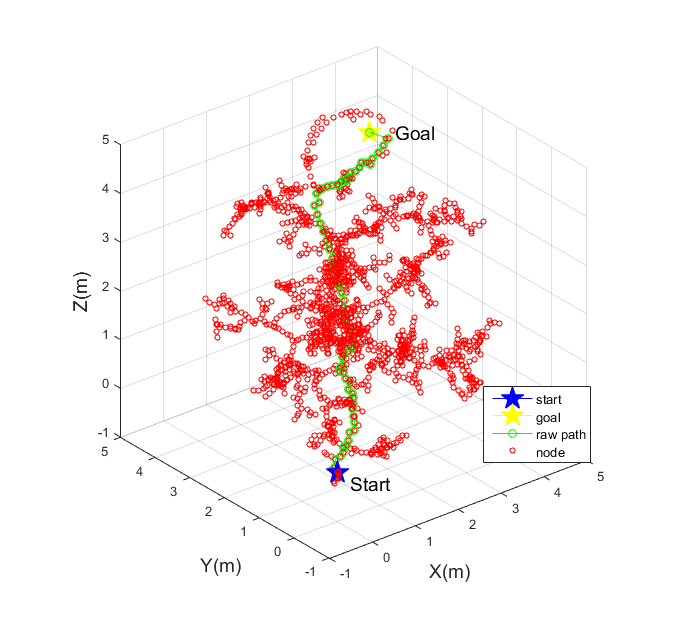}
	\caption{Non-obstacle RRTAUVMS searching result}
	\label{fig:RRTAUVMS1}
\end{figure}

\begin{figure}[htbp]
	\centering
	\includegraphics[scale=0.4]{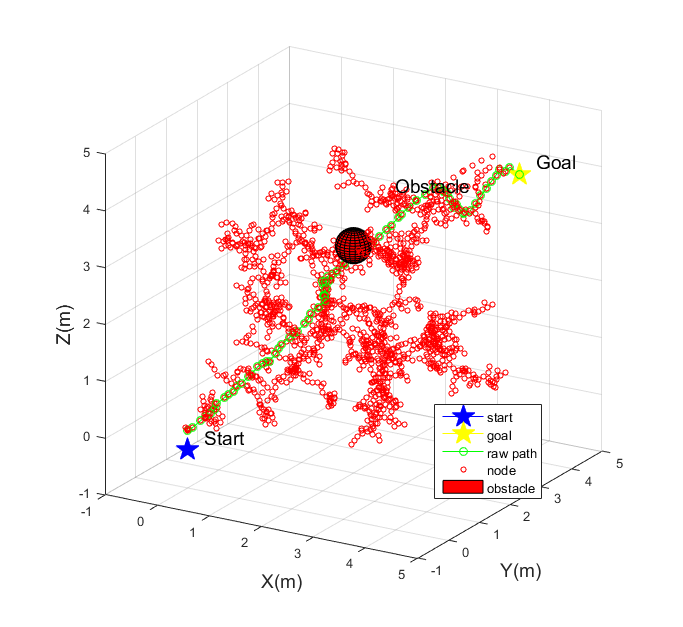}
	\caption{Single obstacle RRTAUVMS searching result}
	\label{fig:RRTAUVMS2}
\end{figure}

\begin{figure}[htbp]
	\centering
	\includegraphics[scale=0.4]{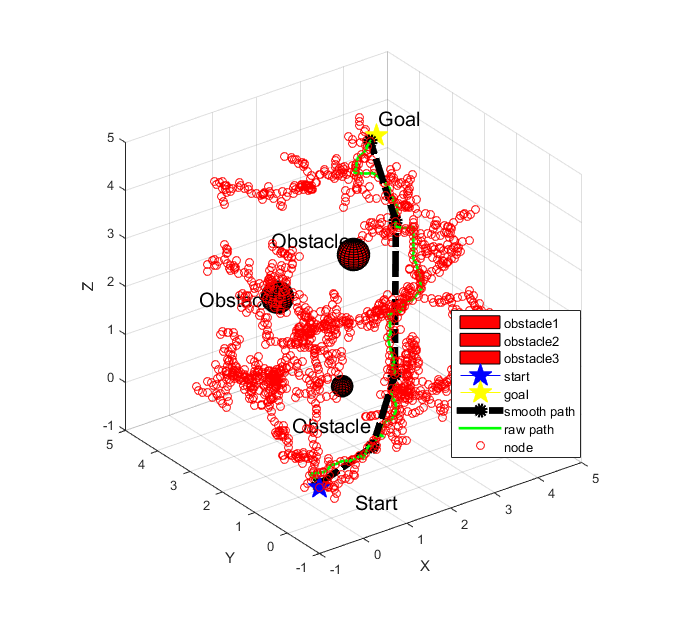}
	\caption{Multi-obstacles RRT searching result}
	\label{fig:RRT_seed_30}
\end{figure}

\begin{figure}[htbp]
      \centering
      \includegraphics[scale=0.4]{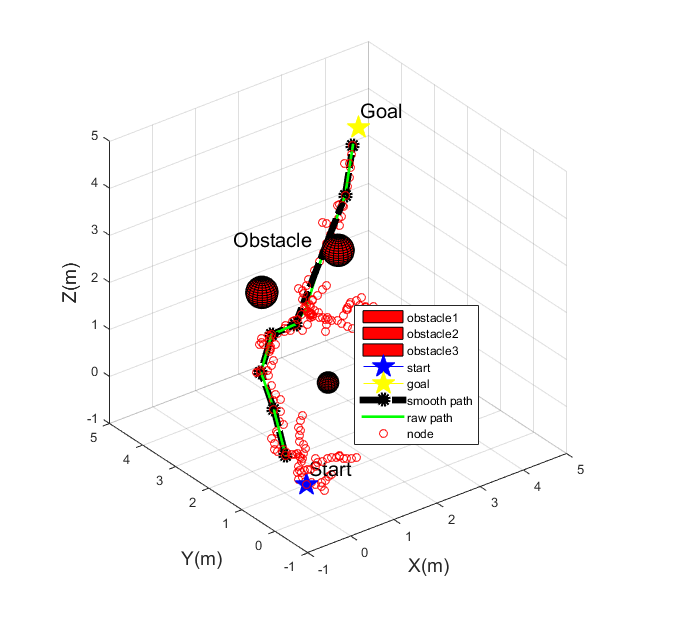}
      \caption{Multi-obstacles RRTAUVMS searching result}
      \label{fig:RRTAUVMS_seed_30}
   \end{figure}

\begin{table}[h]
\caption{Single-Obstacle Path Searching Results}
\label{single_obstacle_path_searching}
\begin{center}
\begin{tabular}{|c||c||c|}
\hline
Seed & RRT searching time(s) & RRTAUVMS searching time(s)\\
\hline
10 & 30.85  & 0.3\\
20 & 18.55  & 0.549\\
30 & 13.77  & 0.3 \\
40 & 2.99   & 0.47\\
50 & 1.39   &0.26\\
\hline
\end{tabular}
\end{center}
\end{table}

\begin{table}[h]
\caption{Multi-Obstacles Path Searching Results}
\label{multi_obstacles_path_searching}
\begin{center}
\begin{tabular}{|c||c||c|}
\hline
Seed & RRT searching time(s) & RRTAUVMS searching time(s)\\
\hline
10 & Not Found  & 0.58\\
20 & 11.3  & 1.91    \\
30 & 26.56  & 1.64 \\
40 & 2.01   & 2.26\\
50 & Not Found   & 1.07\\
\hline
\end{tabular}
\end{center}
\end{table}

The simulation results are shown in Fig.\ref{fig:RRTAUVMS1} $\sim$ Fig.\ref{fig:RRTAUVMS_seed_30}. In these diagrams, the red circles represent the arm tip position for every valid sampling node; The green lines are the final searching results, while the black dash lines are the final smooth paths; Start and goal positions of end effector are represent by the blue pentagram and yellow pentagram. The Fig.\ref{fig:RRTAUVMS1} is the RRTAUVMS algorithm searching result after 3000 times iterations, no obstacle is added in the workspace. Fig.\ref{fig:RRTAUVMS2} is the searching result with one obstacle. These simulations verify the validity of the RRTAUVMS, the algorithm can successfully find a solution whether there is obstacle or not.  Fig.\ref{fig:RRT_seed_30} and Fig.\ref{fig:RRTAUVMS_seed_30} are the comparison studies of the traditional RRT and new proposed RRTAUVMS algorithm. Three obstacles are added to the workspace to test the algorithm performance in complex environment. The charts Fig.\ref{fig:RRT_seed_30} and Fig.\ref{fig:RRTAUVMS_seed_30} show all the collision free sampling nodes when the algorithm first find the satisfactory path. As shown in the charts, the traditional RRT algorithm will run much more iterations than the RRTAUVMS algorithm, so more red circles can be observed in Fig.\ref{fig:RRT_seed_30}. To obtain the quantitative performance of two algorithms, a great deal of simulation experiments are conducted, the comparison results are given in Table.\ref{single_obstacle_path_searching} and Table.\ref{multi_obstacles_path_searching}, the seed values are the MATLAB random number generator input values. The tables record the time consumed by each algorithm with different seed. The efficiency advantage of the proposed algorithm is shown in the table. Whether for single obstacle or multi-obstacles, the planning time of proposed algorithm for single obstacle environment is less than 1 second, the time consumed by the traditional RRT is much more than 1 second for most of the cases. For multi-obstacles environment, the planning time is less than 2 seconds for RRTAUVMS, however, the RRT algorithm can't even find a path in some cases.

\begin{figure}[htbp]
  \centering
  \includegraphics[scale=0.45]{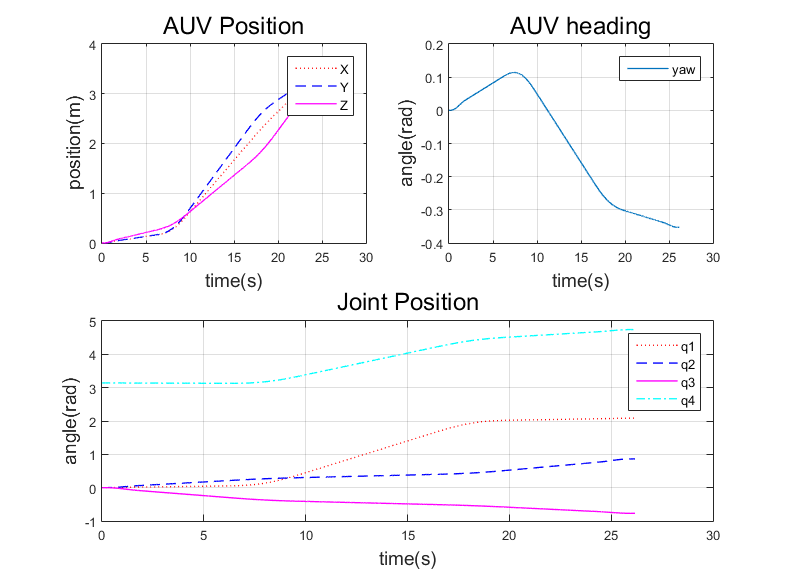}
  \caption{Trajectory RRTAUVMS}
  \label{fig:RRTAUVMSa}
\end{figure}

\begin{figure}[htbp]
  \centering
  \includegraphics[scale=0.45]{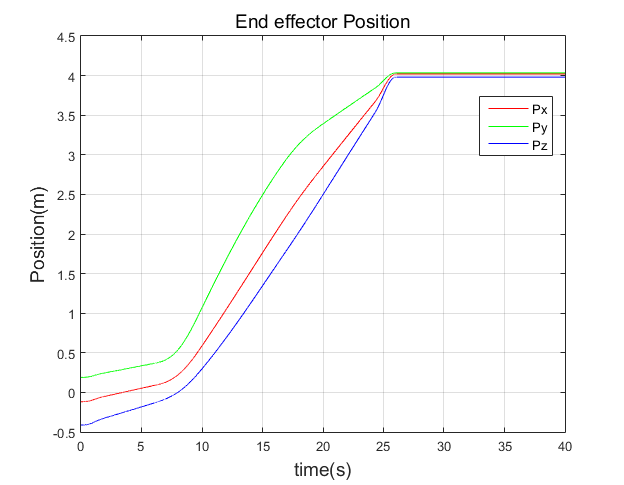}
  \caption{Time histories of end effector position in workspace}
  \label{fig:RRTAUVMSee}
\end{figure}
 
 Fig.\ref{fig:RRTAUVMSa} is the AUVMS trajectories in CSpace, the trajectories are obtained by cubic spline interpolation method with the smoothed RRTAUVMS path, Fig.\ref{fig:RRTAUVMSee} is the time histories of end effector position in workspace. Noting that in Fig.\ref{fig:RRTAUVMSa}, the joint position $q_1, q_2, q_3$ are all in the joint limits, the $q_4$ is a continuous revolution joint, and the $q_4$ position will not affect the end effector position, so it's not bounded to a certain range. From these to diagrams, we can conclude that the proposed algorithm and the path smooth algorithm can give a piratical path for AUVMS system planning. 

\section{CONCLUSIONS}
In this paper, a new path planning algorithm based on RRT and system kinematics is proposed. The AUVMS is a high redundant system, the proposed algorithm could take advantage of the kinematics model to reduce the sampling dimensions, and the joint position constrain is considered by introducing the weighted pseudo-inverse solution. The simulation experiments are conducted after the algorithm is derived.  The simulation results show that the RRTAUVMS algorithm could find a collision free path in complex environment, also it's more efficient than traditional RRT algorithm with a less than 2 seconds searching time.

\addtolength{\textheight}{-12cm}   % This command serves to balance the column lengths

%%%%%%%%%%%%%%%%%%%%%%%%%%%%%%%%%%%%%%%%%%%%%%%%%%%%%%%%%%%%%%%%%%%%%%%%%%%%%%%%

%%%%%%%%%%%%%%%%%%%%%%%%%%%%%%%%%%%%%%%%%%%%%%%%%%%%%%%%%%%%%%%%%%%%%%%%%%%%%%%%



%\section*{ACKNOWLEDGMENT}
% The preferred spelling of the word ÒacknowledgmentÓ in America is without an ÒeÓ after the ÒgÓ. Avoid the stilted expression, ÒOne of us (R. B. G.) thanks . . .Ó  Instead, try ÒR. B. G. thanksÓ. Put sponsor acknowledgments in the unnumbered footnote on the first page.

%%%%%%%%%%%%%%%%%%%%%%%%%%%%%%%%%%%%%%%%%%%%%%%%%%%%%%%%%%%%%%%%%%%%%%%%%%%%%%%%

\bibliography{IEEEabrv,mycitation}
\end{document}